\documentclass{ecmtb2002}

\usepackage[dvips]{epsfig}
\usepackage{exscale}
\begin{document}

\title*{Magnification 
Laws
of Winner-Relaxing 
and Winner-Enhancing Kohonen 
Feature
Maps
\vspace*{-3mm}
 \footnote{Published in: 
%Proc.{} ECMTB 2002, ed.{} V.{} Carpasso et.{} al., MIRIAM series, Milano
V. Capasso (Ed.): Mathematical Modeling \& Computing in Biology and Medicine, 
p. 17--22,
Miriam Series, Progetto Leonardo, Bologna (2003).
}
}
\toctitle{Winner-Relaxing Kohonen Maps}

\titlerunning{Winner-Relaxing Kohonen Maps}

\author{Jens Christian Claussen
%\inst{1}
%\and
% name surname author2
%\inst{2}
}
\authorrunning{J. C. Claussen}
\institute
{Institut f\"ur Theoretische Physik und Astrophysik,
Leibnizstra\ss{}e 15, 24098 Kiel, Germany
(claussen@theo-physik.uni-kiel.de)
%\and  institution 2
\vspace*{-3mm}
 }
\maketitle              % typesets the title of the contribution

%\begin{document}

\index{Claussen, {J. C.}}

\newcommand{\IGN}[1]{ {} }
\setcounter{page}{17}

\pagestyle{myheadings}
\markboth{~~~ \rm J.\ C.\ Claussen \hfill}{\hfill \rm Winner-Relaxing Kohonen Maps ~~~}
\begin{abstract}
Self-Organizing Maps are models for 
unsupervised representation formation 
of cortical receptor fields
by stimuli-driven self-organization
in laterally coupled
winner-take-all feedforward structures.
This paper discusses modifications
of the original Kohonen model 
that were motivated by a potential 
function, in their ability to 
set up a neural mapping of
maximal mutual information.
Enhancing the winner update, instead of relaxing it, 
results in an algorithm that
generates an infomax map 
corresponding to magnification exponent of one. 
Despite there may be more than one algorithm 
showing the same magnification exponent,
the magnification law is an experimentally
accessible quantity and therefore
suitable for quantitative description 
of neural optimization principles.
\end{abstract}
\vspace*{-3mm}

Self-Organizing Maps are one of most successful paradigms in
mathematical modelling of special aspects of brain function, 
despite that a quantitative understanding of the neurobiological 
learning dynamics and its implications on the mathematical process of
structure formation are still lacking.
A biological discrimination between models may be difficult, and
it is not completely clear 
\cite{plumbley}
what optimization goals are dominant 
in the biological development for e.g. skin, auditory, olfactory or retina
receptor fields. 
All of them roughly show a self-organizing ordering
as can most simply be described by the Self-Organizing Feature Map 
\cite{kohonen82} defined as follows:

Every stimulus in input space (receptor field) is assigned to a 
so-called winner (or center of excitation) $\vec{s}$
where the distance $|\vec{v}-\vec{w}_{\vec{s}}|$ to the stimulus is minimal.
According to Kohonen, all weight vectors are updated by
\begin{eqnarray}
\delta \vec{w}_{\vec{r}}=\eta g_{\vec{r}\vec{s}} \cdot 
(\vec{v}-\vec{w}_{\vec{r}}).
\end{eqnarray}
This can be interpreted as a Hebbian learning rule;
$\eta$ is the learning rate,
and $g_{\vec{r}\vec{s}}$ determines the
(pre-defined) topology of the neural layer.
While Kohonen 
chose 
$g$ to be 1 for 
a fixed neighbourhood, and 0 elsewhere,
%% usage of 
a Gaussian kernel (with a width decreasing in time) is more common.
The Self-Organizing Map concept can be used with regular lattices
of any dimension (although 1, 2 or 3 dimensions are preferred 
for easy visualization), with an irregular lattice, none 
(Vector Quantization) 
\cite{vq}, 
or a neural gas
\cite{martinetz_gas} where the coefficients $g$ are 
determined by rank of distance in input space.
In all cases, learning can be implemented by serial (stochastic), batch, or 
parallel updates.

\clearpage
\section{The potential function for the discrete-stimulus non-border-crossing
case}
\begin{figure} \centering
\vspace*{-2mm}
\includegraphics[height=5cm,angle=0]{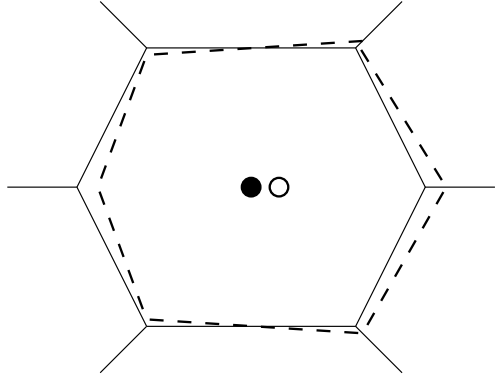}
\vspace*{-1mm}
\caption{Movement of Voronoi borders due to the change of a weight vector
$\vec{w}_{\vec{r}}$.} \label{clfig1}
\end{figure}

One disadvantage of the original Kohonen learning rule
is that it has no potential (or objective, or cost) function
valid for the general case of continuously distributed
input spaces.
Ritter, Martinetz and Schulten \cite{rms91}
gave for the expectation value of the learning step
($F_{\vec{s}}$ denotes the voronoi cell of $\vec{s}$)
\begin{eqnarray}
\langle \delta \vec{w}_{\vec{r}}\rangle
&=&
\eta 
\sum_\mu p(\vec{v}^\mu) g^\gamma_{\vec{r}\vec{s}(\vec{v}^\mu)} 
(\vec{v}^\mu - \vec{w}_{\vec{r}})
\\
&=&
\nonumber
\eta \sum_{\vec{s}} 
g^\gamma_{\vec{r}\vec{s}}
 \sum_{\mu | \vec{v}^{\mu} \in F_{\vec{s}} (\{\vec{w}\})} 
p(\vec{v}^\mu) 
(\vec{v}^\mu - \vec{w}_{\vec{r}})
\stackrel{\mbox{\normalsize ?}}{=} -\eta\nabla V(\{\vec{w}\})
\end{eqnarray}
the following potential function
\begin{eqnarray}
\label{cl_wrkpot}
V_{\sf WRK} (\{\vec{w}\}) = \frac{1}{2} \sum_{\vec{r}\vec{s}} 
g^\gamma_{\vec{r}\vec{s}(\vec{v}^\mu)}
\sum_{\mu | \vec{v}^{\mu} {\in} F_{\vec{s}} (\{\vec{w}\})}
p(\vec{v}^\mu) 
(\vec{v}^\mu - \vec{w}_{\vec{r}})^2.
\end{eqnarray}
This expression is however only 
a valid potential in cases like the end-phase of learning
of Travelling-Salesman-type optimization procedures, i.e.
input spaces with a discrete input probability distribution
and only as long as the borders of the voronoi tesselation
$F_{\vec{s}} (\{\vec{w}\})$ are not shifting across a stimulus vector
(Fig. \ref{clfig1}), which results in discontinuities.

As Kohonen pointed out  \cite{kohonen91}, 
for differentiation 
of (\ref{cl_wrkpot})
w.r. to a stimulus vector $\vec{w}_{\vec{r}}$,
one has to take into account the movement of the borders of the 
voronoi tesselation, leading to corrections to the oringinal
SOM rule by an additional term for the winner;
so (\ref{cl_wrkpot}) is a potential function for the 
Winner Relaxing Kohonen 
%(WRK) 
rather than for  SOM.
This approach has been
generalized 
\cite{claussen_nc}
to obtain infomax maps,
as will be discussed in Section \ref{clau_sec4}.

\clearpage
\section{Limiting cases of WRK potential and Elastic Net}
For a gaussian neighbourhood kernel
\begin{eqnarray}
g^\gamma_{\vec{r}\vec{s}}
\sim \exp(- {(\vec{r}-\vec{s})^2}/{2\gamma^2}),
\end{eqnarray}
it is 
%meaningful 
illustrative
to look at limiting cases for the kernel width $\gamma$.
The limit of $\gamma\to\infty$ implies $g_{\vec{r}\vec{s}}$ to be constant,
which means that all neurons receive the same learning step and
there is no adaptation at all.
On the other hand, 
the limit $\gamma=0$ gives $g^\gamma_{\vec{r}\vec{s}}
=\delta_{\vec{r}\vec{s}} $
which coincides
with the so-called Vector Quantization 
(VQ)
 \cite{vq} which means 
%that
 there is no neighbourhood interaction at all.

The interesting case is where $\gamma$ is small, 
which corresponds to the parameter choice for the
end phase of learning.
Defining $\kappa := \exp(-1/2\gamma^2)$,
 we can expand 
($\kappa$-expansion of the Kohonen algorithm
\cite{claussen_dipl})
\begin{eqnarray}
g^\gamma_{{r}{s}}
= \delta_{{r}{s}}
+ \kappa (\delta_{r s-1} + \delta_{r s+1} ) + o(\kappa^2).
\end{eqnarray}
Here we have written the sum over the two next neighbours 
in the one-dimensional case (neural chain), but the
generalization to higher dimensions is straightforward.
(Note that instead of evaluating the gaussian for each
learning step, one saves a considerable amount of 
computation by storing the
powers of $\kappa$ in a lookup table for a moderate
kernel size, and neglecting the small contributions outside.
Using the $(\kappa,1,\kappa)$ kernel instead of the
``original'' $(1,1,1)$ Kohonen kernel 
reduces fluctuations
and preserves the
magnification law better; 
see \cite{ritter91} for the corrections for a non-gaussian kernel.)

If we now restrict to a Travelling Salesman setup
(periodic boundary 1D-chain) with the case that the number of neurons
equals the number of stimuli (cities), 
then the potential reduces to
\begin{eqnarray}
V_{\sf WRK;TSP} =  \frac{1}{2} \sum_{\mu}
|\vec{v}^\mu - \vec{w}_s|^2 
+ \kappa \sum_r |\vec{w}_{r+1}-\vec{w}_{r}|^2 + o(\kappa^2).
\end{eqnarray}
This coincides with the $\sigma\to 0$ limit 
(the limit of high input resolution, or low temperature)
of the Durbin and Willshaw Elastic Net
\cite{durbin87} 
which also has a local (universal) magnification law 
(see section \ref{clau_sec3})
in the 1D case \cite{claussen_icann}, that however astonishingly 
(it seems to be the only feature map where it is no power law)
is not a power law; low magnification is 
delimited by elasticity).
As the Elastic Net is troublesome concerning 
parameter choice \cite{simmen91}
for stable behaviour esp.{} for
serial presentation (as a feature map model would require),
the connection between Elastic Net and WRK
should be taken more as a motivation to study the WRK map.
Apart from ordering times and reconstruction errors, 
one quantitative measure for feedforward structures is
the transferred mutual information, 
which is related to the magnification law,  
as described in the following Section. 
% \ref{clau_sec3}.

\clearpage
\section{The Infomax principle and the Magnification Law}
\label{clau_sec3}
Information theory \cite{shannon}
gives a quantitative framework to describe
information transfer through feedforward systems.
Mutual Information between input and output is 
maximized when output and input are always identical, 
then the mutual information is maximal and
equal to the information entropy of the input.
If the output is completely random, the mutual information
vanishes.
In noisy systems, maximization of mutual information
{e.g.} leads to the optimal number of redundant transmissions.

Linsker \cite{linsker89} 
was the first who applied this ``infomax principle''
to a self-organizing map
architecture (one should note he used a slightly different
formulation of the neural dynamics, 
and the algorithm itself is computationally very costly).
However, the approach can be used to quantify
information preservation even for other algorithms.

This can be done in a straightforward manner by
looking at the magnification behavoiur of a self-organizing map.
The magnification factor is defined as the
number of neurons (per unit volume in input space).
This density is equal to the inverse of the
Jacobian of the input-output mapping.
The remarkable property of many self-organizing maps is that
the magnification factor (at least in 1D) 
becomes  a function of the local input probability 
density (and is therefore independent of the desity elsewhere,
apart from the normalization),
and in most cases even follows a power law.
While the Self-Organizing Map shows a power law with exponent 
$2/3$ \cite{ritter86},
an exponent of 1 would correspond (for a map without added noise)
to maximal mutual information, or on the other hand to 
the case that the firing probability of all neurons is the same.

\begin{figure} 
\centering
\vspace*{-4.5mm}
\includegraphics[height=5.9cm,angle=0]{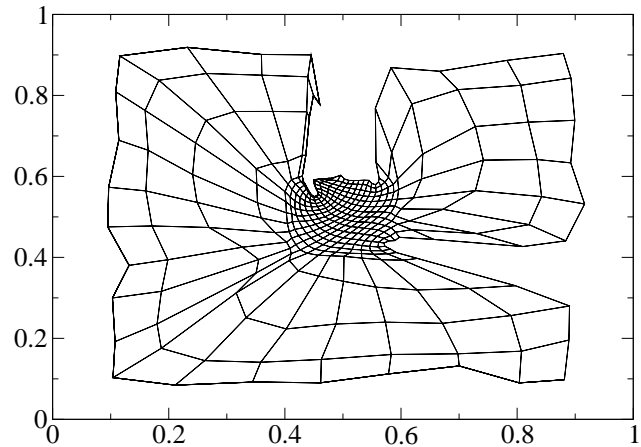}
\vspace*{-1mm}
\caption{For $D\geq{}2$ networks, the neural density
 in general is not a local function of the stimulus density,
but depends also on the density in neighbouring regions.
} \label{clfig2}     
\end{figure}

In higher dimensions, however, the 
stationary state of the weight vectors
will in general not decouple, so the 
magnification law is no longer 
only of local dependence of the 
input probability density (Fig. \ref{clfig2}).

\clearpage
\section{Generalized Winner Relaxing Kohonen Algorithms}
\label{clau_sec4}
As pointed out in \cite{claussen_nc}, the prefactor $1/2$ in
the WRK learning rule can be replaced by a free parameter, 
giving the Generalized Winner Relaxing Kohonen Algorithm
\begin{eqnarray}
\delta \vec{w}_{\vec{r}} =
\eta
\Big(
g^\gamma_{\vec{r}\vec{s}}
 (\vec{v}^\mu - \vec{w}_{\vec{r}}) 
- \lambda
\delta_{\vec{r}\vec{s}}
\sum_{\vec{r}^{'}\neq \vec{s}}
g^\gamma_{\vec{r}^{'}\vec{s}}
(\vec{v} - \vec{w}_{\vec{r}^{'}})
\Big).
\end{eqnarray}
The $\delta_{\vec{r}\vec{s}}$ restricts the modification to the 
winner update only (the first term is the classical Kohonen SOM).
The subtracted ($\lambda>0$,
winner relaxing case)
resp.{} added ($\lambda<0$, winner enhancing case)
sum corresponds to some center-of-mass movement
of the rest of the weight vectors.

\begin{figure}
\centering
\includegraphics[height=5.7cm]{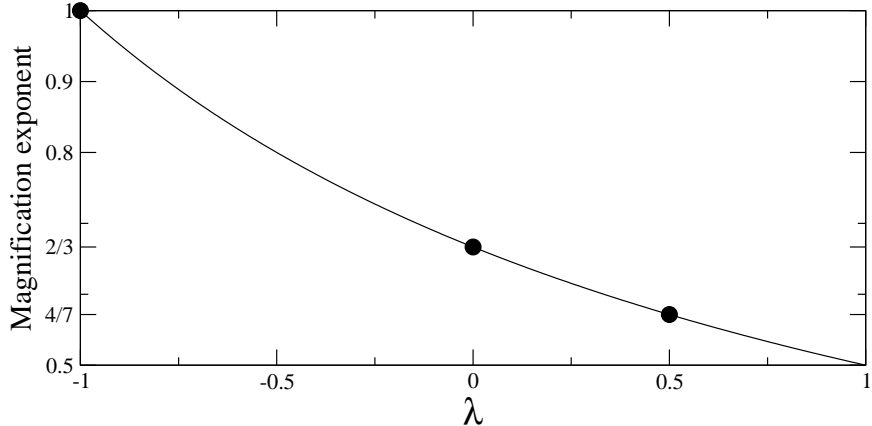}
\caption{The Magnification exponent of the 
Generalized Winner Relaxing Kohonen 
as function of $\lambda$, 
including the special cases
SOM ($\lambda=0$) and WRK ($\lambda=1/2$). 
}
\label{clfig3}       % Give a unique label

\end{figure}

The Magnification law has been shown to be a power law
\cite{claussen_nc} with exponent $4/7$ for the Winner
Relaxing Kohonen (associated with the potential) and
$2/(3+\lambda)$ for the Generalized Winner Relaxing Kohonen
(see Fig. \ref{clfig3}).
As stability for serial update can be acheived only within
$-1 \leq \lambda \leq +1$, 
the magnification exponent can be adjusted by an  {\sl a priori}
choice of parameter $\lambda$ between $1/2$ and $1$.

Although Kohonen reported the  WRK
to have a larger fraction of initial conditions ordered after finite time
 \cite{kohonen91}, 
one can on the other hand ask how fast 
a rough ordering is reached from a random
initial configuration.
Here the average ordering time can have a minimum 
\cite{claussen_delmenhorst}
for negative 
$\lambda$ 
corresponding to a near-infomax regime.
As in many optimization problems, this seems 
mainly to be a question whether 
the average, the maximal (worst case),
or the minimal (parallel evaluation)  
ordering time 
is to be minimized.

\clearpage
\section{Call for Experiments and Outlook}
Cortical receptor fields of an adult animal show plasticity on 
a long time-scale which is very well separated
from the time-scale of the signal processing
dynamics. 
Therefore, for constant input space distributions
(which in principle can be measured) the magnification
law could be accessed experimentally by
measuring the neural activity (or the number of active
neurons) by any electrical or noninvasive technique.
Especially the auditory cortex is well suitable for 
a direct comparison of mathematical modelling 
due to its 1-D input space. 
Owls and bats have a large amplitude variation in 
their input probability distribution 
(their own ``echo'' frequency is heard most often)
and are therefore pronounced candidates for experiments.

In a refined step, experimental and theoretical 
investigations on the nonlinear modifications of the
learning rules have to be done.
On the experimental side, it has to be 
clarified which refinements to long-term-potentiation
and long-term-depletion have to be found for 
the weight vectors in a neural map architecture.
Mechanisms that can be included as modifications are
modified winner updates (as for GWRK), 
probabilistic winner selection
\cite{graepel97,heskes99energyfunctions},
or a local learning rate, depending on 
averaged firing rates and reconstruction errors
\cite{herrmann95,villmann}.
On the theoretical side, it is obvious that 
the same magnification exponent can be obtained by
quite different algorithms.
The relation 
between them and the transfer to more realistic
models should be investigated further.

% Non-BibTeX users please use

%%%%%%%%%%%%%%%%%%%%%%%%%%%%%%%%%%%%%%%%%%%%%%%%%%%%%%%%%%%%%%%%%%%%%%

%
\end{document}